# Hierarchical Solution of Markov Decision Processes using Macro-actions

Milos Hauskrecht, Nicolas Meuleau
Leslie Pack Kaelbling, Thomas Dean
Computer Science Department, Box 1910
Brown University, Providence, RI 02912
{*milos, nm, lpk, tld*}@cs.brown.edu

Craig Boutilier
Department of Computer Science
University of British Columbia
Vancouver, BC V6T 1Z4, Canada
*cebly@cs.ubc.ca*

## Abstract

We investigate the use of temporally abstract actions, or macro-actions, in the solution of Markov decision processes. Unlike current models that combine both primitive actions and macro-actions and leave the state space unchanged, we propose a hierarchical model (using an *abstract MDP*) that works with macro-actions only, and that significantly reduces the size of the state space. This is achieved by treating macro-actions as local policies that act in certain regions of state space, and by restricting states in the abstract MDP to those at the boundaries of regions. The abstract MDP approximates the original and can be solved more efficiently. We discuss several ways in which macro-actions can be generated to ensure good solution quality. Finally, we consider ways in which macro-actions can be reused to solve multiple, related MDPs; and we show that this can justify the computational overhead of macro-action generation.

## 1 Introduction

Markov decision processes (MDPs) [11, 22] have proven tremendously useful as models of stochastic planning and decision problems. However, traditional dynamic programming remains computationally intractable for practical problems, requiring time polynomial in the size of the state and action spaces, but where these spaces are generally too large to be explicitly enumerated. Considerable research has been directed toward the solution of Markov decision processes (MDPs) with large state and action spaces. These include function approximation [2], reachability analyses [5] and aggregation techniques [7, 3, 4].

Despite these advances, little attention has been paid to the *reuse* of policies or value functions generated for one MDP in the solution of a related MDP. While such reasoning is common in classical planning—for instance, through the use of *macros* [8, 16, 13] or plan repair strategies [15]—its application in stochastic settings is less common. Suitable techniques of this type could lead to the amortization of solution costs over a large number of problems, and the ability to solve future problem instances quickly, which is critical to on-line reasoning.

One of the few models to deal with solution reuse within the MDP framework is the Skills model of Thrun and Schwartz [24], which attempts to learn how to reuse policy fragments (or *skills*) for different tasks. Another is found in the work of Sutton and his colleagues [23, 20, 21], who have developed models of *macro-actions* for MDPs that can be reused to solve multiple MDPs when objectives (or goals) change. In particular, macros are viewed as "local" policies that are implemented until some termination condition is met, at which point a new macro (or any other action) can be applied. Key to the success of this framework is the ability to construct models of macro-actions that allow them to be treated as if they were ordinary actions in the original MDP.

In this paper, we continue the investigation of the use of macros in MDPs; specifically, we focus on the problem of planning with macro-actions addressed by Precup, Sutton and Singh [21]. Our main aim is the development of a different model for planning with macros that deals with some of the computational problems associated with this earlier model (the *PSS model*). In particular, while the PSS model allows macros designed for one MDP to be applied to a related MDP, it still relies on explicit dynamic programming over the state space of the related MDP (and a larger action space). Thus it does nothing to alleviate the problem of large state spaces. Furthermore, the use of macros is not guaranteed to reduce the time required to find an optimal solution.

We present in Section 2 a *hierarchical model* for the use of macro-actions that specifically addresses the difficulties of large state and action spaces. We take a macro to be a local policy, defined over a region of state space, that terminates when that region is left. We show how an *abstract MDP* can be constructed that consists only of states that lie on the borders of adjacent regions, and whose solution determines a policy that consists of macros *only*. Hierarchical models similar to the one we propose have been investigated by Forestier and Varaiya [9], and recently by Parr [18, 19].

Two limitations of this model are then addressed. The first relates to solution quality. Since the policy generated by solving the abstract MDP can contain only macros, certain behaviors cannot be realized, thus the resulting policy may be suboptimal. In Section 3, we identify conditions under which the set of macros that comprise the action space of the abstract MDP give rise to an $\varepsilon$-optimal policy for the original MDP. We then consider both systematic and heuristic techniques for macro generation that ensure high quality behavior.

The second limitation relates to solution time, specifically the time needed to generate a set of good macros. This generally requires that we perform some form of dynamic programming within specific regions of the state space. Since our regions cover the state space and macros should capture a variety of control behaviors in different regions, macro generation can become computationally more intensive than solving the original MDP.[1] This problem can be diminished if we can generate macros off-line for fast on-line reasoning, or reuse them to solve multiple problems. In Section 4, we briefly analyze the requirements for feasible macro reuse and describe a hybrid model in which changes in the original MDP (either in the reward function or the system dynamics) lead to an expansion of some parts of the abstract MDP, which can then be solved. Since this *hybrid MDP* consists primarily of abstract states and macro-actions, it can be solved effectively and provide for fast, on-line response to changes in problem specification. The use of macros for the on-line solution of multiple related MDPs is the main advantage of our hierarchical model.

## 2 A Hierarchical Model of Macro-actions

### 2.1 Markov Decision Processes

A (finite) *Markov decision process* is a tuple $\langle S, A, T, R \rangle$ where: $S$ is a finite set of states; $A$ is a finite set of actions; $T$ is a transition distribution $T : S \times A \times S \to [0, 1]$, such that $T(s, a, \cdot)$ is a probability distribution over $S$ for any $s \in S$ and $a \in A$; and $R : S \times A \to \mathbf{R}$ is a bounded reward function. Intuitively, $T(s, a, w)$ denotes the probability of moving to state $w$ when action $a$ is performed at state $s$, while $R(s, a)$ is the immediate reward associated with action $a$ in $s$.

Given an MDP, the objective is to construct a *policy* that maximizes expected accumulated reward over some horizon of interest. We focus on *infinite horizon, discounted* decision problems, where we adopt a policy that maximizes $E(\sum_{t=0}^{\infty} \beta^t \cdot r^t)$, where $r^t$ is a reward obtained at time $t$ and $0 < \beta < 1$ is a discount factor. In such a setting, we restrict our attention to stationary policies of the form $\pi : S \to A$, with $\pi(s)$ denoting the action to be executed in state $s$. The value of a policy $\pi$ can be shown to satisfy [11]

$$V_\pi(s) = R(s, \pi(s)) + \beta \sum_{t \in S} T(s, \pi(s), t) \cdot V_\pi(t).$$

A policy $\pi$ is *optimal* if $V_\pi(s) \geq V_{\pi'}(s)$ for all $s \in S$ and policies $\pi'$. The *optimal value function* $V^*$ is the value function for any optimal policy.

A number of techniques for constructing optimal policies exist. An especially simple algorithm is *value iteration* [1]. We produce a sequence of value functions $V^n$ by starting from an arbitrary $V^0$, and defining

$$V^{i+1}(s) = \max_{a \in A} \{R(s, a) + \beta \sum_{t \in S} T(s, a, t) \cdot V^i(t)\}. \quad (1)$$

The sequence of functions $V^i$ converges to $V^*$ in the limit. Each iteration is known as a *Bellman backup*. After some finite number $n$ of iterations, the choice of maximizing action for each $s$ forms an optimal policy $\pi$ and $V^n$ approximates its value.

### 2.2 Macro-actions and their Models

Sutton [23] has argued that it is crucial to be able to model MDPs at multiple time scales. The ability to determine the value—within an underlying MDP—of a complex sequence of actions or program is important in, say for example, robot programming. In the navigation problem illustrated in Figure 1(a), a programmer may have provided a program (or partial policy) that enables the robot to exit one of the rooms through a particular door. Integrating such a partial policy into the decision process is a difficult task given that: (a) the robot usually "commits" to the execution of this program; and (b) the program extends over some period of time. To deal with this problem, Precup, Sutton and Singh [23, 20, 21] have developed *multi-time models* and applied them to planning with MDPs. In what follows, we draw heavily on the use of these multi-time models. Parr and Russell [17] have proposed a related model in which a (partial) policy is modeled using a finite-state machine. These policies are then "abstracted" hierarchically and treated as primitive actions to be invoked by higher-level behaviors.

While such *temporally abstract actions*, or *macro-actions*, are useful for modeling constrained behavior—such as partially specified policies—we also view them as a useful tool that allows the *reuse* of a solution generated for one MDP in the solution of another. However, this perspective casts macros in a very different light. While Sutton and his colleagues have not explicitly considered how macros arise, we focus on the issues associated with the automatic generation of macro actions. Rather than supposing a temporally abstract behavior has been provided, we imagine that the decision maker will be forced to deal with a number of related problem-solving episodes, and desires a set of macros that will help solve these MDPs more quickly. Thus the effort required to generate these macros (something not considered in the PSS model) will "pay for itself" either with decreased reaction time to changing circumstances, or with total computational savings over multiple problem instances. The Skills model of Thrun and Schwartz [24] has a similar motivation, though they do not address the use of multi-time models for learned skills. Parr [18, 19] has independently investigated the use of hierarchical models with an eye toward macro generation, and has considered many of the same problems we address here.

---

[1] We note that aggregation and approximation techniques can be used within a region, though we do not address this issue here.

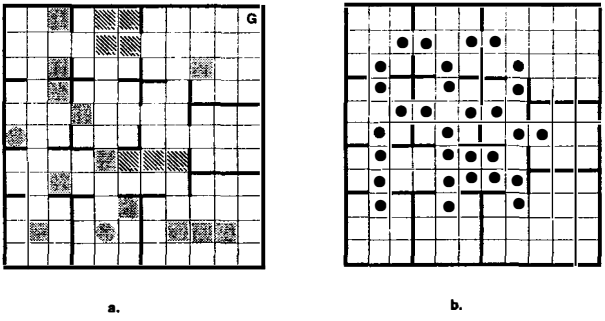

Figure 1: (a) Test problem Maze 121. Shaded squares denote locations with higher cost, patterned squares represent areas in which moves are more uncertain (a move in the intended direction is less likely). Shaded circles denote absorbing states with a finite positive cost, G stands for a zero cost goal state; (b) peripheral states for the partitioning into 11 rooms (regions).

Formally, our model relies on a *region-based decomposition* of a given MDP $\langle S, A, T, R \rangle$ as defined by Dean and Lin [6].

**Definition 1** *A region-based decomposition $\Pi$ of an MDP $M = \langle S, A, T, R \rangle$ is a partitioning $\Pi = \{S_1, \cdots, S_n\}$ of the state space $S$. We call the elements $S_i$ of $\Pi$ the regions of $M$. For any region $S_i$, the exit periphery of $S_i$ is*

$$XPer(S_i) = \{t \in S - S_i : T(s, a, t) > 0 \text{ for some } a, s \in S_i\}.$$

*The entrance periphery of $S_i$ is*

$$EPer(S_i) = \{t \in S_i : T(s, a, t) > 0 \text{ for some } a, s \in S - S_i\}.$$

*We call elements of $XPer(S_i)$ exit states for $S_i$ and elements of $EPer(S_i)$ entrance states. The collection of all peripheral states is denoted*

$$Per_\Pi(S) = \cup_i \{EPer(S_i) : i \leq n\} = \cup_i \{XPer(S_i) : i \leq n\}.$$

Figure 1(b) shows the set of peripheral states obtained if we partition the problem of Figure 1(a) into the eleven regions corresponding to different rooms.

A *macro-action* is simply a local policy defined for a particular region $S_i$. Intuitively, this policy can be executed whenever an agent is in the region and terminates when the agent leaves the region (if ever).

**Definition 2** *A* macro-action *for region $S_i$ is a local policy $\pi_i : S_i \to A$.*

Our definition is much more specific than that of PSS, who define macros using arbitrary starting and termination conditions, and allow mappings that depend on the time elapsed or the trajectory followed since the macro action was initiated. Within our framework, the starting condition for macro $\pi_i$ would simply be $s^t \in S_i$ (we are in the region) and the termination condition would be $s^t \notin S_i$ (we are out of the region).

A key insight of PSS (which finds its roots in earlier work by Sutton [23]) is that one can treat a macro-action of this type as a *primitive action* in the original MDP if one has an appropriate reward and transition model for the macro. They propose the following method of modeling macros.

**Definition 3** *A discounted transition model $T_i(\cdot, \pi_i, \cdot)$ for macro $\pi_i$ (defined on region $S_i$) is a mapping $T_i : S_i \times XPer(S_i) \to [0, 1]$ such that*

$$\begin{aligned} T_i(s, \pi_i, s') &= E_\tau(\beta^{\tau-1} \cdot \Pr(s^\tau = s' \mid s^0 = s, \pi_i)), \\ &= \sum_{t=1}^{\infty} \beta^{t-1} \cdot \Pr\left(\tau = t, s^t = s' \mid s^0 = s, \pi_i\right) \end{aligned}$$

*where the expectation is taken with respect to time $\tau$ of termination of $\pi_i$. A discounted reward model $R_i(\cdot, \pi_i)$ for $\pi_i$ is a mapping $R_i : S_i \to \mathbf{R}$ s.t.*

$$R_i(s, \pi_i) = E_\tau(\sum_{t=0}^{\tau} \beta^t R(s^t, \pi_i(s^t)) \mid s^0 = s, \pi_i),$$

*where the expectation is taken with respect to completion time $\tau$ of $\pi_i$.*

The discounted transition model is a standard stochastic transition matrix specifying the probability of leaving $S_i$ via a specific exit state given that $\pi_i$ was initiated at a specific state inside the region, with one exception: the probability is *discounted* according to the expected time at which that exit occurs. As demonstrated by PSS, this clever addition allows the transition model to be used as a normal transition matrix in any standard MDP solution technique, such as policy or value iteration.[2] The discounted reward model is similar, simply measuring the expected accrued reward during execution of $\pi_i$ starting from a particular state in $S_i$.

---

[2]Our definition of the discounted transition model differs slightly from that of PSS: their transition model is obtained by

## 2.3 Constructing Macro Models

Since we are concerned with the automatic generation of macros, we now consider the construction of discounted transition and reward models for macros. Issues related to the macro-model construction are discussed also in [19].

Let $\pi_i$ be a macro defined on $S_i$. The discounted transition probability $T_i(s, \pi_i, s')$ for $s \in S_i$, macro $\pi_i$ and $s' \in XPer(S_i)$ satisfies:

$$T_i(s, \pi_i, s') = \\ T(s, \pi_i(s), s') + \beta \sum_{s'' \in S_i} T(s, \pi_i(s), s'') T_i(s'', \pi_i, s').$$

This leads to $|XPer(S_i)|$ systems of linear equations, one set for every exit state. Each system consists of $|S_i|$ equations with $|S_i|$ unknowns. The systems can be solved either directly or using iterative methods. Thus, the time complexity of finding all transition probability parameters is $O(|XPer(S_i)||S_i|^3)$.

We can construct the reward model in a similar fashion. Let $R_i(s, \pi_i)$ be the expected discounted reward for following the policy $\pi_i$ starting at state $s \in S_i$. Then we have:

$$R_i(s, \pi_i) = R(s, \pi_i(s)) + \beta \sum_{s' \in S_i} T(s, \pi_i(s), s') R_i(s', \pi_i).$$

This defines a set of $|S_i|$ linear equations, which can be solved in $O(|S_i|^3)$ time.

Overall, the computation of macro parameters takes $O((|XPer(S_i)| + 1)|S_i|^3)$ time per macro. Note that an overhead for generating macros (finding suitable policies defining macros and computing their parameters) may become, in many instances, computationally more expensive than solving the original MDP problem. Thus we must carefully consider what kinds of planning situations justify the computational effort.

## 2.4 The Hierarchical Solution of MDPs with Macros

Suppose we are given an MDP $M$, a decomposition $\Pi$, and a set of macros $A_i = \{\pi_i^1, \cdots, \pi_i^{n_i}\}$ for each region $S_i$ associated with this partition. There are two reasonably direct ways in which these can be used to solve $M$ more efficiently.

First, we can simply add these macro-actions to $M$; let $M_a$ denote the *augmented MDP* constructed by extending the action space from $A$ to $A \cup A_1 \cup \cdots \cup A_n$, assuming that macro models are used to determine transitions and rewards associated with these new actions. $M_a$ can be solved

---

multiplying our variable $T_i$ by the constant $\beta$. Our definition is consistent with the update formula in Equation 1, while PSS use a formula where the discount factor is folded into the transition model (this requires multiplying the one-step transition probabilities by $\beta$ before using them).

by standard methods, such as value iteration. Because all *base level* actions (those in $A$) are present, the policy so constructed is guaranteed to be optimal. Furthermore, the presence of macros can enhance the convergence of value iteration, as demonstrated by Sutton et al. [20, 21]. This is due to the fact that the single "application" of a macro can propagate values through a large number of states and over a large period of time in a single step. In general, this model requires more work per iteration because of the increased action space, but potentially fewer iterations. We note that this savings does not account for the overhead associated with generating macros and constructing the models for each macro.

We also note that, depending on the initial value function used to begin value iteration, macros can actually *increase* the number of steps required for convergence compared to the value iteration with primitive actions alone. Specifically, Hauskrecht [10] showed that if $V^0$ is an upper bound on the optimal value function $V^*$, value iteration in the augmented MDP is guaranteed to require at least as many iterations as in the original MDP (the same holds for a lower bound and minimization of costs). An empirical demonstration of this phenomenon is provided in next section.

Alternatively, we can imagine a *reduced MDP*, $M_r$, formed by replacing $A$ with $A_1 \cup \cdots \cup A_n$. $M_r$ will generally be more efficiently solvable because there are fewer actions to consider, and convergence will be enhanced as above. However, because the possible behaviors one can consider are limited to the application of these macros, there is no guarantee that the resulting solution is optimal: this will depend crucially on the macros introduced.

While these models offer some advantages, they do not use macros to alleviate the problem of state space size. Each method requires explicit value iteration over the state space, with possibly a larger number of actions. We instead wish to solve a much smaller MDP, taking advantage of the fact that, by committing to the execution of a macro, decisions need only be made at peripheral states, not at states that lie strictly within a region. To capture this intuition, we consider the *hierarchical* application of macro operators within a high-level, or *abstract*, MDP. This model is closely related to the "landmark" technique developed by Kaelbling [12] for learning policies for hierarchical stochastic shortest path problems.

**Definition 4** *Let* $\Pi = \{S_1, \cdots, S_n\}$ *be a decomposition of MDP* $M = \langle S, A, T, R \rangle$, *and let* $\mathbf{A} = \{A_i : i \leq n\}$ *be a collection of macro-action sets, where* $A_i = \{\pi_i^1, \cdots, \pi_i^{n_i}\}$ *is a set of macros for region* $S_i$. *The abstract MDP* $M' = \langle S', A', T', R' \rangle$ *induced by* $\Pi$ *and* $\mathbf{A}$, *is given by:*

- $S' = Per_\Pi(S) = \cup \{EPer(S_i) : i \leq n\}$;

- $A' = \cup_i A_i$ *with* $\pi_i^k \in A_i$ *feasible only at states* $s \in EPer(S_i)$;

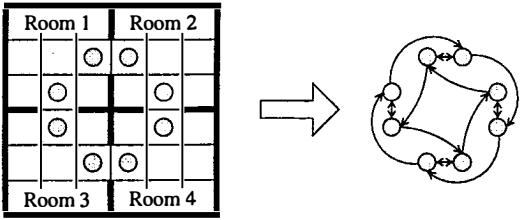

Figure 2: Abstract MDP for a four-room example. Grey circles mark peripheral states of the original MDP, i.e. states of the abstract MDP.

- $T'(s, \pi_i^k, t)$ is given by the discounted transition model for $\pi_i^k$, for any $s \in EPer(S_i)$ and $t \in XPer(S_i)$; $T'(s, \pi_i^k, t) = 0$ for any $t \notin XPer(S_i)$;

- $R'(s, \pi_i^k)$ is given by the discounted reward model for $\pi_i^k$, for any $s \in EPer(S_i)$.

The transition and reward models required by the abstract MDP are restricted to peripheral states and make no mention of states "internal" to a region. Due to discounting in $T'$ these definitions do not describe an MDP, but they do preserve the Markov property;[3] thus, we may use dynamic programming techniques to solve the abstract MDP. An abstract MDP for a simple four-room navigation problem is shown in Figure 2. Regions are formed by the rooms and the peripheral states make up the abstract MDP. We assume macros exist that can take the robot out of any room through any door, accounting for the connectivity of the abstract MDP.

Notice that the abstract MDP induced by a given decomposition can be substantially smaller than the original MDP, especially if the problem can be decomposed into a number of regions with relatively small peripheries—this is the case in our running example, and in the types of domains considered in [20, 21].

We call a policy $\pi': S' \to A'$ for $M'$ that maps peripheral states to macro-actions a *macro-policy*. Such a policy $\pi'$, when considered in the context of the original MDP $M$, defines a *non-Markovian* policy $\pi$; that is, the choice of action at a state $s$ can depend on previous history. In particular, the action $\pi(s)$ to be executed at some state $s \in S_i$ will generally depend on the state $s_e$ by which $S_i$ was most recently entered: $\pi(s) = \pi'(s_e)(s)$.[4]

---

[3] Specifically, the probability of moving from any entrance state to an exit state for a given macro is independent of previous history.

[4] Note that the macro-policy does not dictate the actions to take if the process begins in an internal state $s$ of some region $S_i$. To deal with this, we can use a greedy macro choice with regard to the "intermediate macro models," $R_i$ and $T_i$, and the values of the abstract MDP at $XPer(S_i)$. This is required only for the initial state, all other decisions are made at peripheral states in the abstract MDP. Note that the greedy approach can be applied also to generate a markovian policy for all states in the region (see [10] for details).

Once a set of macros has been provided, along with their models, our hierarchical approach induces a problem with a considerably smaller state space (and often a smaller action space). This computational advantage comes at a price however—the possibility of generating a suboptimal policy. This is due to the fact that the abstract MDP allows the decision maker to consider only a limited range of behaviors. Therefore it is important to ensure that the macros provided (or generated) offer a choice of behaviors that are of acceptable value. We will turn our attention to this issue in Section 3.

### 2.5 Experimental results

To demonstrate the computational savings made possible by our hierarchical approach to planning with macros, we have performed experiments on the simple navigation problem in Figure 1. The agent can move in any compass direction to an adjacent cell or stay in place. The move actions are stochastic, so the agent can move in an unintended direction with some small probability. The objective is to minimize the expected discounted cost incurred by navigating the maze, with each state, except the zero-cost absorbing goal state, incurring some cost. The costs and transition probabilities are not uniform across the maze.

We compared the results of value iteration for the original MDP, the augmented MDP and the abstract MDP, the latter two formed using the rooms in the problem as regions. The macros were formed heuristically using the simple strategy described in Section 3.2, giving $|XPer(S_i)| + 1$ macros for every region $S_i$. Figure 3 shows how the estimated value (minimal expected cost) of a particular state improves with the time (in seconds) taken by value iteration on each of the three models. When the initial value function estimate is an upper bound, both the augmented MDP and the abstract MDP lead to faster convergence of the value function. In the augmented MDP, the ability of macros to propagate value through a large number of states produces large changes in the value function in a single iteration step, overcoming the increased number of actions. Note, however, that when the initial estimate of the value function is a lower bound, the augmented MDP actually performs worse than the original MDP. These effects would be reversed if we were maximizing rewards instead of minimizing costs. The abstract MDP has significantly reduced state and action spaces sizes. Although in general, macros can lead to suboptimal value functions (and subsequently policies), in our example, the abstract MDP produced nearly optimal policies (and did so very quickly). The average time (in seconds) taken per value iteration step in this example is 0.045 for the original MDP, 0.12 for the augmented MDP, and 0.019 for the abstract MDP. This reflects the increased action space of the augmented MDP and the reduced action

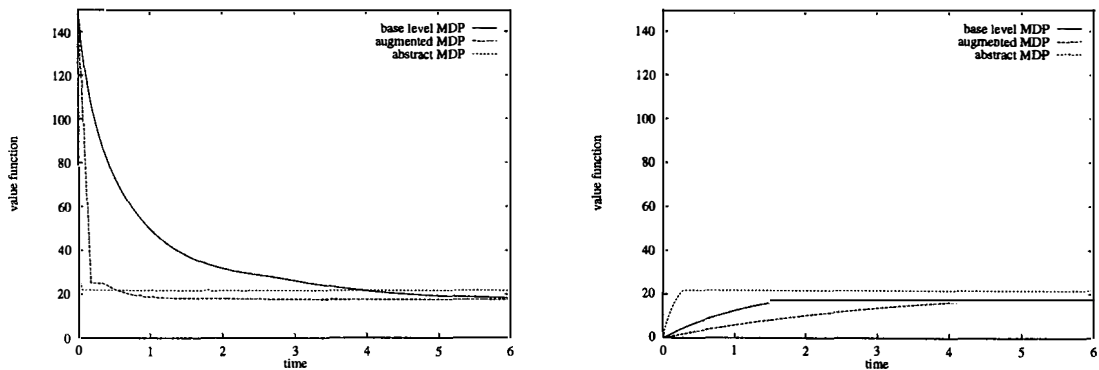

Figure 3: Solution quality versus time for various models. Results using a properly initialized value function (w.r.t. the augmented MDP) are shown on the left. Results for a poor initial function are shown on the right. In the latter case, the augmented MDP converges more slowly than the original MDP.

and state spaces for the abstract MDP, as expected.

## 3 Construction and Quality of a Macro Set

While macros can speed up computation, the question remains just how good the resulting policies will be. In particular, within our hierarchical model, the space of policies that can be considered is severely restricted. Thus, we wish to ensure that the macros used admit the "flexibility" of behavior needed to discover good policies. The problem is less pressing for augmented MDPs—since base actions are available, optimality is assured—but still important if convergence is to be enhanced.

A primary goal of a macro-selection strategy is to find a small set of good macros, that is, macros that are likely to produce, when combined, a good approximation of the optimal solution.

### 3.1 Macro Generation using Peripheral Values

Suppose we offer the robot in our running example two macros for possible execution in Room 1 of Figure 2, each corresponding to a policy that attempts to leave the room by one of the two exits. We are making an implict assumption that one of these two behaviors is desirable, and thus that there is no good reason to hang around in that room. We may prescribe rather different local policies for the room containing the goal; there is a reason not to leave the region.

This suggests a general way to automatically generate macro actions for region $S_i$. We want to trade off the rewards associated with the states in $S_i$ with the values of leaving the region via some exit state. This tradeoff is naturally modeled and analyzed as a *local MDP* where rewards are attached to states in $S_i$ and estimated *values* are attached to elements of its exit periphery.

**Definition 5** *Let $S_i$ be a region of MDP $M =$*

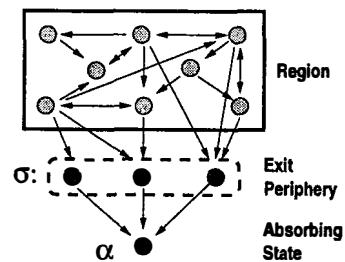

Figure 4: A Local MDP for Macro Generation

$\langle S, A, T, R \rangle$, *and let $\sigma : XPer(S_i) \to R$ be a seed function for $S_i$. The local MDP $M_i(\sigma)$ associated with $S_i$ and $\sigma$ consists of: (a) state space $S_i \cup XPer(S_i) \cup \{\alpha\}$; (b) actions, dynamics and rewards associated with $S_i$ as in $M$; (c) a reward $\sigma(s)$ associated with each $s \in XPer(S_i)$; and (d) a single cost-free action applicable at each $s \in XPer(S_i)$ that leads with certainty to $\alpha$ (a cost-free absorbing state).*

The local MDP is depicted graphically in Figure 4. Solving $M_i(\sigma)$ results in a local policy $\pi_i$ whose behavior is optimal if the seed function $\sigma$ reflects the true value of reaching specific exit states.

Intuitively, if we could seed the exit periphery of each local MDP using a function $\sigma$ within $\varepsilon$ of the true value function at these states, we could generate a single macro for each region, and "string them together" to obtain an approximately optimal policy. More precisely, we have:

**Theorem 1** *Let $\Pi = \{S_1, \cdots, S_n\}$ be a decomposition of MDP $M$, and let $V$ be the optimal value function for $M$. Let $A = \cup \{A_i : i \leq n\}$ be a set of macro actions such that each $A_i$ contains some macro $\pi_i$ generated by the local MDP $M_i(\sigma_i)$ where $|\sigma_i(s) - V(s)| \leq \varepsilon$ for all $s \in XPer(S_i)$. If $M'$ is the abstract MDP induced by $\Pi$ using action set $A$, and $V'$ is the optimal value function for $M'$,*

*then*

$$|V'(s) - V(s)| \leq \frac{2\varepsilon\beta}{1-\beta}$$

*for all $s \in S'$ (the abstract state space). Furthermore, if $\tau$ is a lower bound on the completion time of all macros, then*

$$|V'(s) - V(s)| \leq \frac{2\varepsilon\beta^\tau}{1-\beta^\tau}.$$

Note that more precise error bounds can be found when "effective" discounting rates are considered for every macro transition.

### 3.2 Construction of Macro Sets

The previous result indicates that knowledge of the (optimal) value function for an MDP can give rise to good macros. Of course, such prescience is rare: if we knew the value function, we would have no decision problem to solve. However, we often have heuristic knowledge regarding the range of the value function at certain states, or constraints on its possible values. It is precisely this type of knowledge that comes into play when one imposes partial policies (say, in the form of a control routine). Even *some* information can be used to construct a good set of macros that guarantees approximately optimal performance. We consider several methods for exploiting such knowledge.

If one knows the *range* of the value function, this can be used to construct a set of macros systematically. For instance, when constructing macros for Room 1 in Figure 2, suppose our knowledge of the value function is sparse—all we know is that the values of the two exit states lie between $V_{\min}$ and $V_{\max}$.[5] In order to generate a set of macros for Room 1 that is guaranteed to contain a good macro, we can use the *coverage technique*: intuitively, for each of the two exit states, we consider values that lie in the range $[V_{\min}, V_{\max}]$ spaced some $\delta$ apart; that is, we consider a grid or mesh covering $[V_{\min}, V_{\max}]^2$. By constructing macros for each $\sigma$ lying on a grid point, we are assured that one such $\sigma$ is within $\frac{1}{2}\delta$ of the optimal value function and that (assuming other regions have "good" macros from which to choose) close-to-optimal behavior results when the abstract MDP is solved.

This coverage technique can be extremely expensive: given such generic knowledge of the value function, we will generate $[(V_{\max} - V_{\min})/\delta)]^{|XPer(S_i)|}$ macros per region. However, we can often do much better. First, the number of macros is usually smaller than the number of grid points covering $[V_{\min}, V_{\max}]$. Thus it is often more appropriate to search a local policy space. One technique for doing so was suggested recently by Parr [18]. Second, we can apply various forms of domain-specific knowledge. For instance, the values of several exit states for a region $S_i$ may not be known, but we may know that these values are (approximately) the same (e.g., they are equidistant from any rewarding or dangerous states). This effectively reduces the dimensionality of the required grid. Tighter constraints on the value function can reduce the range of values that need to be tried. Furthermore, in circumstances where no reward can be obtained within the region, only differences in the *relative values* of exit states impact the local policy: this too can reduce the number of macros needed.

---

[5]Such bounds are easily obtainable using the maximum and minimum rewards.

The systematic coverage technique can lead to a generation of a large number of macros per region. Thus, unless tight constraints are known on the value function, this can involve substantial overhead and, in many instances, be unprofitable. *Heuristic* methods for macro generation can alleviate this difficulty if they require the construction of a small number of macros. One such strategy, suggested by Sutton et al. [21] for robot navigation problems such as our example, involves creating macros for each region $S_i$ that try to lead the agent out of $S_i$ via different exit states. To do so requires seeding a local MDP such that one exit state gets high value and all others get low value. We described experiments with this heuristic technique in the previous section, but we also added a *stay-in-region* macro that keeps the agent in the region, by seeding all exit states with low values. This technique leads to a set of $|XPer(S_i)| + 1$ macros per region.

In general, the above heuristic strategy assures that exits and potential goals within the region will not be overlooked while planning at the abstract level. Note, however, that this technique does not guarantee that the necessary coverage will be obtained. For example, while implementing a policy to exit in one way, the agent may find itself actually "slipping" closer to another exit due to uncertainty in its actions. However, the policy will ensure the agent persists in its attempt to leave as planned. If both exit states have equal value, forcing the agent to choose one or the other can be far from optimal. Instead, we would like to use a third macro that takes the agent to the *nearest* exit. However, we cannot discard the original macros unless we know *in advance* that the values are similar. In addition, unless one accounts for potential variability in the actual value assigned to an exit state, sound decisions to stay within a region or leave it cannot be made.

Finally, we mention the possibility of using iterative refinement techniques for macro construction. A simple refinement strategy uses the value function produced by solving the abstract MDP as seeds for an entirely new set of macros. In particular, we choose an initial set of seeds, generate a single macro per region, then solve the induced abstract MDP. The resulting value function is used as a seed to generate a new set of macros (again one per region), and the new abstract MDP is solved. This iterative macro-refinement method is a special case of asynchronous policy iteration [2] and is similar to Dantzig-Wolfe (D-W) decom-position techniques [6, 14]. D-W techniques can be viewed as iterative schemes for evaluating and modifying macro sets generated by assigning values to peripheral states.

In general, iterative macro-refinement methods overcome the threat of poor initial seeding (and the generation of poor macros) by gradually improving the macro set using information as it becomes available. These approaches require the repeated construction of new macros, which may limit their applicability. We leave deeper investigation of iterative techniques for future work.

## 4  Multiple MDPs and the Reuse of Macros

### 4.1  Hybrid MDPs

As discussed above, generating macro-actions and constructing their transition and reward models is an intensive process, requiring explicit state space enumeration. If a large number of macros is generated, the overhead associated with this process will outweigh any speed-up provided by macros during value iteration. Thus our hierarchical approach (or any approach requiring macro model generation) may not be worthwhile as a technique to solve a single MDP.

The main reason to incur the overhead of macro construction lies in the reuse of macros to solve multiple related MDPs. In our running example, the robot may have constructed a policy that gets it to the goal consistently, but at some point the goal location might change, or the penalties associated with other locations may be revised, or perhaps the environment (or its abilities) changes so that the uncertainty associated with its moves at particular locations increases. Any of these changes requires the solution of a new MDP, reflecting a change in reward structure or change in system dynamics. However, the changes to the MDP are often *local*: the reward function and the dynamics remain the same in all but a few regions of state space. For instance, it may be that the goal location moves within Room 3, but no other part of the reward function changes.

Local changes in MDP structure can induce global changes in the value function (and can induce dramatic qualitative changes in the optimal behavior). If macros have been generated for a region such that they cover a set of different behaviors, they can be applied and reused in solving these revised MDPs. However, there is one impediment to the application of macroactions to revised MDPs, namely, the fact that revising an MDP requires that the local information (rewards or dynamics) for some region must change. In our example, the macros for most regions can be reused; but those generated for Room 3 do not reflect the revisions in reward or transition probabilities described above. One possibility would be to generate new macros for revised regions. However, this could lead to computational inefficiencies and delays as discussed earlier. Instead, it is often

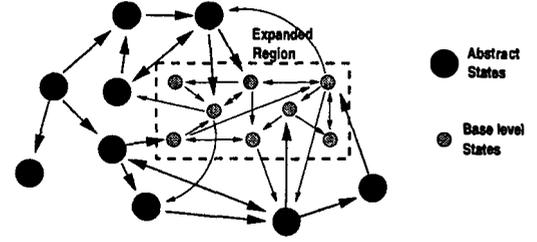

Figure 5: A Hybrid MDP

easier to solve revised MDPs using a *hybrid MDP*, containing both abstract and base level states.

**Definition 6** *Let* $\Pi = \{S_1, \cdots, S_n\}$ *be a decomposition of MDP* $M = \langle S, A, T, R \rangle$, *and let* $M' = \langle S', A', T', R' \rangle$ *be the abstract MDP induced by* $\Pi$ *and macro set* $\mathbf{A} = \{A_i : i \leq n\}$. *Let* $\overline{M} = \langle S, A, \overline{T}, \overline{R} \rangle$ *be a* local revision *of* $M$ *with regard to region* $S_i$; *that is,* $T(s,a,t) = \overline{T}(s,a,t)$ *and* $R(s,a) = \overline{R}(s,a)$ *for all* $s \notin S_i$. *The* hybrid expansion $M^* = \langle S^*, A^*, T^*, R^* \rangle$ *of* $M'$ *by* $\overline{M}$ *is:*

- $S^* = Per_\Pi(S) \cup S_i$;

- $A^* = \cup\{A_j \in \mathbf{A} : j \neq i\} \cup A$, *where* $\pi_j^k \in A_j$ *is feasible only at states* $s \in EPer(S_j)$, *and* $a \in A$ *is feasible only at* $S_i$;

- $T^*(s, \pi_j^k, t)$ *is given by the discounted transition model for* $\pi_j^k$, *for any* $s \in EPer(S_j)$ *and* $t \in XPer(S_j)$ $(j \neq i)$; $T^*(s, \pi_j^k, t) = 0$ *for any* $t \notin XPer(S_j)$; $T^*(s,a,t) = \overline{T}(s,a,t)$ *for any* $s \in S_i$ *and* $t \in S^*$;

- $R^*(s, \pi_j^k)$ *is given by the discounted reward model for* $\pi_j^k$ *for any* $s \in S_j (j \neq i)$, *while* $R^*(s,a) = \overline{R}(s,a)$ *for any* $s \in S_i$.

Thus the hybrid MDP $M^*$, constructed when the structure within region $S_i$ changes, consists of the original abstract MDP with the abstract states in $EPer(S_i)$ replaced by the region $S_i$ itself. This is depicted graphically in Figure 5. We note that this expansion is easily defined for changes in any number of regions.

While there may be substantial overhead in creating macros, these can be reused to solve multiple problems, thus amortizing the cost over a number of problem-solving episodes. More importantly, the use of hybrid MDPs has considerable advantages when real-time response is required to changing circumstances. Given a new MDP $M^k$ that differs from a base MDP $M^0$ in a single region $S_i$ (or, more generally, some small set of regions), this new problem can be solved using a hybrid MDP of size $|S'| + |S_i - EPer(S_i)|$ (recall $S'$ is the set of peripheral states, or states in the abstract MDP). For example, if an MDP is partitioned into $k$ regions of roughly uniform size, and the average size of the entrance periphery of any region is $p$, then a hybrid MDP with one expanded region has roughly $kp + \frac{|S|}{k}$ states. Without the use of macros and abstract/hybrid MDPs, the solution of a new problem requires value or policy iteration over the entire state space of size $|S|$. Thus a new problem can be solved much more quickly. The off-line generation of macros can lead to very efficient on-line solution of new problem instances.

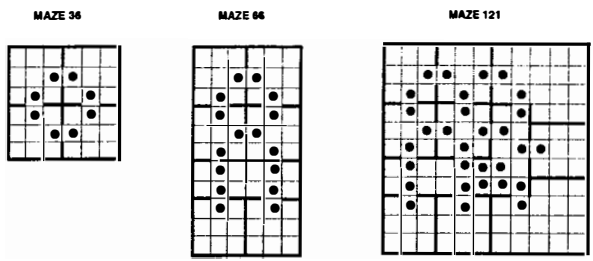

Figure 6: Problems used to test the benefits of macro-reuse. Circles denote peripheral states assumed by the hybrid-MDP method.

### 4.2 Experimental results

To illustrate the potential for speed-up in on-line response time for multiple related MDPs using macro actions and our hybrid MDP model, we compared response time of value iteration for both the base level MDP and the hybrid MDP on three sequences of related problems. We examined three robot navigation problems of increasing complexity, shown in Figure 6: Maze 36 with 36 states and 4 regions; Maze66 with 66 states and 7 regions; and Maze 121 with 121 states and 11 regions. In each instance, the underlying MDP was modified locally by changing the goal, represented by a zero cost absorbing state (this required changes to both the dynamics and reward model).

Table 1 summarizes results obtained for 25 problem instances (using different randomly selected goal states) and two value iteration methods working with the base level MDP and the hybrid MDP. A heuristic set of macro-actions, described in Section 3.2, was used for the hybrid MDP. Value iteration was started using the solution obtained for the original (locally unmodified) MDP and stopped when a fixed precision (0.01 cost units) was achieved.

The results illustrate that the hybrid MDP model, given suitable macros, can solve new problem instances much more quickly than resolving the MDP with the original state and action spaces. We also see that the savings offered by the hybrid model are greater for larger problems, exactly as expected. This is due to the fact that local changes affect a significantly smaller proportion of the original model. For a hybrid MDP this means that most of the structure of the abstract MDP is preserved and only the regions in which the change has occured are elaborated.

A disadvantage of the hybrid MDP framework is that one has to generate and precompute a set of macros, which can be computationally very costly.[6] However, if the macro construction process is performed in advance (off-line), this delay may be unimportant in relation to the improved ability to solve new problem instances quickly. Alternatively, the delay can be justified when the computational cost could be amortized over multiple problem instances. For example, based on our test results, the hybrid MDP method in this example would start to dominate (in terms of a total solution time, which counts both the delay and time to solve $n$ tasks) after 22, 23, and 24 tasks are solved for Maze 36, Maze 66, and Maze 121, respectively. Notice that amortization threshold (the number of tasks after which macro preparation "pays off") increases slowly with problem size, even though this sequence of problems is such that a more complex maze has roughly double the state space size of its predecessor. This trend seems promising for the application of macros in very large domains with many possible tasks or goals.

The hybrid MDPs used in our experiments rely on a set of heuristically generated macros (see Section 3.2). The macro set is relatively small and performed very well on the set of maze navigation problems we tested. This is documented by comparing AEC scores, measuring average expected cost for all peripheral states and for 25 randomly generated goal tasks. The increase in the cost score for larger problems is caused by an increase in distances between peripheral and possible goal states. The practical creation of good macro sets for different types of problems remains an interesting open issue.

## 5 Conclusions

We have proposed a new hierarchical model for solving MDPs using macro actions. Our *abstract MDP* allows potentially dramatic reductions in the size of state and action spaces. This requires commitment to the execution of macro actions—they cannot be reconsidered at each stage—thus leading to potentially inflexible, suboptimal behavior. We have elaborated conditions and macro construction techniques that provide guarantees on solution quality. Within this model, anytime tradeoffs can be made rather easily. Furthermore, with *hybrid MDPs*, we have a technique that allows macros to be reused to solve multiple MDPs, providing for fast, on-line decision making, and allowing macro construction costs to be amortized over many problem solving episodes.

There are a number of questions and open issues that remain to be addressed within this framework and many interesting directions in which this work can be extended. We have ignored the question of where partitionings of state space come from: apart from handcrafted decompositions, one can imagine several strategies for automatic decomposition. However, there are several dimensions along which partitionings can be compared: larger regions often lead to smaller peripheries, which result in smaller abstract MDPs (which in turn can be solved more readily), and increase the odds that a revision of the MDP will be localized to a small number of regions; smaller regions, in contrast, allow macros to be generated more quickly when revisions are required and often lead to smaller hybrid MDPs (fewer base states are added to the expanded MDP). These trade-offs need to be addressed in a systematic fashion.

---

[6]We note that approximation techniques can be used to alleviate this problem.

|  | Maze 36 | | | Maze 66 | | | Maze 121 | | |
| --- | --- | --- | --- | --- | --- | --- | --- | --- | --- |
|  | delay | av.time | AEC | delay | av.time | AEC | delay | av.time | AEC |
| base MDP | 0 | 1.22 | 5.96 | 0 | 2.61 | 8.55 | 0 | 5.94 | 9.96 |
| hybrid MDP | 5.52 | 0.96 | 6.01 | 12.62 | 2.04 | 9.73 | 24.47 | 4.89 | 10.72 |

Table 1: Results obtained by base and hybrid MDP methods on 25 randomly selected goals and three navigation problems. The delay (in seconds) measures the time spent to prepare macroactions, av. time is the average time (in seconds) to converge to a solution of required precision (0.01), AEC measures the solution quality, and is computed by averaging expected cost over all peripheral states and task instances.

Other interesting questions surround the use of concise MDP representations (e.g., Bayes nets) to form decompositions and to solve local, abstract and hybrid MDPs. Related is the need to concisely represent macros and macro models without explicit enumeration of the state space.

The reuse of macros naturally suggests an extension of the analysis provided here, and the questions posed above, to deal with known distributions over problem instances. If we have information pertaining to the ways in which system dynamics and reward functions may be revised, we would like to exploit it in forming our decomposition of state space and the macros one provides.

## Acknowledgements

We would like to thank Ronald Parr for a motivating discussion on macro-actions and for pointing out additional references. This work was supported in part by DARPA/Rome Labs Planning Initiative grant F30602-95-1-0020 and in parts by NSF grants IRI-9453383 and IRI-9312395. Craig Boutilier was supported by NSERC Research Grant OGP0121843 and IRIS-II Project IC-7, and this work was undertaken while the author was visiting Brown University. Thanks also to the generous support of the Killam Foundation.